\documentclass[lettersize,journal]{IEEEtran}
\usepackage{amsmath,amsfonts,color}
\usepackage{algorithmic}
\usepackage{algorithm}
\usepackage{array}
\newcolumntype{R}[1]{>{\raggedleft\arraybackslash}p{#1}}
\usepackage[caption=false,font=normalsize,labelfont=sf,textfont=sf]{subfig}
\usepackage{textcomp}
\usepackage{stfloats}
\usepackage{url}
\usepackage{verbatim}
\usepackage{graphicx}
\usepackage{cite}
\usepackage{booktabs}
\usepackage{cuted}
\usepackage{capt-of}
\usepackage{pdflscape}
\usepackage{placeins}
\usepackage{afterpage}
\usepackage{adjustbox}
\usepackage{tabularx}
\usepackage{bbding}
\usepackage{balance}

\begin{document}

\title{Trust in LLM-controlled Robotics: a Survey of Security Threats, Defenses and Challenges}
%Attack and Defense for LLM-controlled Robotics}
\author{Xinyu Huang$^*$, Shyam Karthick V B$^*$, Taozhao Chen$^*$, Mitch Bryson, Thomas Chaffey, Huaming Chen\textsuperscript{\Envelope}, Kim-Kwang Raymond Choo\textsuperscript{\Envelope}, Ian R. Manchester\textsuperscript{\Envelope} 
\thanks{Xinyu Huang, Shyam Karthick V B, Taozhao Chen, Thomas Chaffey and Huaming Chen are with the School of Electrical and Computer Engineering, The University of Sydney, Sydney, 2006, Australia. (email: huaming.chen@sydney.edu.au)}
\thanks{Mitch Bryson and Ian R. Manchester are with the Australian Centre for Robotics and School of Aerospace, Mechanical and Mechatronic Engineering, The University of Sydney, Australia. E-mail address: ian.manchester@sydney.edu.au.}
\thanks{Kim-Kwang Raymond Choo is with the Department of Information Systems and Cybersecurity, University of Texas at San Antonio, San Antonio, TX 78249-0631, USA. Part of this research was undertaken at the School of Engineering and Natural Sciences, University of Iceland, 102 Reykjavík, Iceland. (e-mail: raymond.choo@fulbrightmail.org).}
\thanks{$^*$ indicates these authors contributed equally to this work as first authors. \Envelope lists the corresponding authors.}
\thanks{The work of K.-K. R. Choo is supported by the Cloud Technology Endowed Professorship, and the Fulbright Distinguished Scholar in Cybersecurity and Critical Infrastructure award.}
}



\maketitle

\begin{abstract}
The integration of Large Language Models (LLMs) into robotics has revolutionized their ability to interpret complex human commands and execute sophisticated tasks. However, such paradigm shift introduces critical security vulnerabilities stemming from the ``embodiment gap'', a discord between the LLM's abstract reasoning and the physical, context-dependent nature of robotics. While security for text-based LLMs is an active area of research, existing solutions are often insufficient to address the unique threats for the embodied robotic agents, where malicious outputs manifest not merely as harmful text but as dangerous physical actions. In this work, we present a systematic survey, summarizing the emerging threat landscape and corresponding defense strategies for LLM-controlled robotics. Specifically, we discuss a comprehensive taxonomy of attack vectors, covering topics such as jailbreaking, backdoor attacks, and multi-modal prompt injection. In response, we analyze and categorize a range of defense mechanisms, from formal safety specifications and runtime enforcement to multi-LLM oversight and prompt hardening. Furthermore, we review key datasets and benchmarks used to evaluate the robustness of these embodied systems. By synthesizing current research, this work highlights the urgent need for context-aware security solutions and provides a foundational roadmap for the development of safe, secure, and reliable LLM-controlled robotics.
\end{abstract}


\section{Introduction}
\IEEEPARstart{T}{he} integration of Large Language Models (LLMs) into robotic systems marks a transformative step towards more capable and general-purpose autonomous agents \cite{liang2023code,ren2023robots,rana2023sayplan, wang2024llmbasedrobottaskplanning}. By leveraging their extensive world knowledge and reasoning capabilities, LLMs empower robots to interpret high-level human commands, enabling complex tasks execution and intuitive human-robot interaction \cite{guan2023leveraging, tellex2020robots, wu2023multimodal, Wang2024LLMforRobotics, Mon-Williams2025, Wang_2024, sikorski2024deploymentlargelanguagemodels, ding2024robottaskplanningsituation, aghzal2025surveylargelanguagemodels, Arkin2020Multimodal}. Several seminal framework have demonstrated the impacts of this integration. In \cite{ahn2022do}, \textit{SayCan} is proposed to address the grounding problem by merging the LLM's understanding of language with robotic affordances, thereby aligning a robot's semantic planning with actual physical capabilities. Similarly, the \textit{Inner Monologue} approach \cite{huang2023inner} enables a robot to ``think'' in a closed loop with LLM. It allows the system to reason iteratively about environment changes, so that the robot can self-correct and improve its plans for complex, long-term tasks. These solutions have significantly expanded capabilities across key areas such as perception, decision-making, and code generation for low-level control.

However, the deployment of LLMs in robotics introduces a critical and underexplored set of safety challenges. While LLMs excel at abstract reasoning, they are inherently agnostic to the physical context, lacking an intrinsic understanding of metrics, sensor data, or dynamic physics \cite{liu2025aligningcyberspacephysical, raptis2025agentic, wu2025vulnerabilityllmvlmcontrolledrobotics,ouyang2024llmsenseharnessingllmshighlevel, lu2025mindgapdivergencehuman}. This fundamental disconnect is often referred to as the \textit{`embodiment gap'}, creating a new class of vulnerabilities that traditional security measures fail to mitigate. While previous surveys have largely focused on either the capabilities of LLMs for robotics or general LLM security, they don't comprehensively address the unique threats arising when the powerful models control the physical system~\cite{zeng2023large, Kumar2024Adversarial}.

This gap is particularly evident in adversarial scenarios \cite{sadasivan2025attackersnoisemanipulateaudiobased, zhang2024badrobot, liu2024robustness, robey2024jailbreakingllmcontrolledrobots}. While numerous defenses have been developed for text-based LLMs \cite{wu2023defending, zhang2023defending, piet2024jatmo, pisano2023bergeron, kumar2023certifying, cao2023defending, alon2023detectinglanguagemodelattacks, kim2023robust}, they often prove ineffective in the context of LLM-controlled robotics. The core challenge lies in the definition of ``harm'' shifting from semantic toxicity in text to physical interaction in motion. Recent work shows that, a successful jailbreak on a robot requires not only bypassing safety filters but also generating syntactically valid and physically executable commands that can cause real-world damage \cite{robey2024jailbreakingllmcontrolledrobots, karnik2025embodiedredteamingauditing, lu2025poexpolicyexecutablejailbreak}. Furthermore, unlike chatbot defenses that can flag universally malicious content, the safety of a robotic action is inherently state-dependent, i.e., a command to ``move forward'' may be perfectly safe in open space but catastrophic near a cliff or human operator. These are the nuances that traditional LLM safety mechanisms are ill-equipped to handle. A growing body of research highlights that incorporating LLMs in robotics expands the system's attack surface. It necessitates a departure from security research that focuses solely on the trustworthiness of a single, disembodied agent \cite{liu2024compromising,wang2025trojanrobot,wu2025vulnerabilityllmvlmcontrolledrobotics,zhang2024badrobot}.

%\subsection{Our Contributions} Raymond: Should not have a single subsection within a section
To address this gap, this survey fills a critical gap by providing a systematic overview of the emerging security threat landscape and corresponding defense strategies for LLM-controlled robotics. Specifically, we focus on attacks targeting the cognitive layer, the reasoning and planning modules where natural language understanding occurs. It is distinct from the attacks on low-level perception, vision, or motor control modules. By isolating risks that exploit the linguistic intelligence of LLM-controlled robotics, we aim to facilitate a roadmap for future research, highlighting the open challenges and guiding the development of robust and context-aware robotics systems.
%Specifically, we focus on attacks that target the language model and reasoning components, which serves as the cognitive layer for natural language understanding and high-level planning. —rather than attacks on perception, vision, or low-level action modules.By highlighting these novel risks that exploit the linguistic intelligence of LLMs, we hope to provide a roadmap for future research, guiding the development of more robust systems and fostering a community committed to secure and responsible deployment.

Specifically, the key contributions of this work are as follows:
\begin{itemize}
\item \textit{Novel Risks Highlight:} We elucidate the unique threats that emerge at the intersection of linguistic intelligence and physical embodiment in LLM-controlled robotics.
\item \textit{Attack Vector Taxonomy:} We propose a comprehensive taxonomy of attack vectors that translate malicious prompts and data into tangible, real-world harm.
\item \textit{Defenses Analysis:} We provide a structured analysis of current defense strategies and integrated frameworks for threat mitigation.
\item \textit{Research Agenda:} We outline a roadmap for future research to guide the development of robust and secure LLM-controlled robotic systems.
\end{itemize}
%In the next section, we first discuss our survey protocol.

\section{Survey Protocol and Comparison}
With the rapid rise of LLMs, a significant body of survey literature has emerged. A review of this corpus delineates two distinct research trajectories: one exploring the models' inherent risks and security issues, and the other extending their applications in complex real-world environments. Despite the breath of research in both domains, a critical intersection remains underexplored: the unique security vulnerabilities and physical risks that arise by the integration of LLMs into robotic systems. \textit{To date, this specific topic lacks a systematic survey.} This section aims to outline the evolution of existing survey landscape, analyze its influence on LLM-controlled robotics, thereby establishing the novelty and value of our work.

With the seminal work of GPT-3 by Brown et al. \cite{brown2020language}, the first comprehensive survey by Zhao et al. \cite{zhao2023survey} systematically mapped the field, presenting a taxonomy of technical architectures, key training resources, and evaluation frameworks that laid the foundation for following researches. Since then, the research community has advanced along two parallel paths.

The first path focuses on the latent security risks and vulnerabilities of LLMs. Foundational work by Liu et al. \cite{liu2023trustworthy} pioneered the discussion of ``trustworthiness'' as a core issue, while Yao et al. \cite{yao2024survey} further dissected the dual nature of LLMs in the security domain. Subsequent studies such as Shayegani et al. \cite{shayegani2023surveyvulnerabilitieslargelanguage} and Das et al. \cite{das2024security}, established comprehensive attack taxonomies and proposed corresponding countermeasures. However, these works share similar focus on harms, such as data contamination, misleading outputs, or disinformation, that are inherently digital and conceptual. It doesn't provide investigation concerning the physical manifestations of these vulnerabilities.

Simultaneously, the second path has been dedicated to the deployment of LLMs within robotic systems. Early surveys, such as Zeng et al. \cite{zeng2023large}, summarize nascent applications which captured the state of this emerging cross-disciplinary field. Following this work, a group of most recent surveys explored distinct sub-fields, including agent architectures \cite{wang2024survey}, foundation model applications across the full robot autonomy stack \cite{hu2024generalpurposerobotsfoundationmodels,Firoozi2024FoundationRobotics}, manipulation-specific approaches \cite{li2025foundationmodelsbringrobot}, and component-wise integration frameworks \cite{kim2024survey}. These works explored how to design effective structures and methodologies to better leverage LLM capabilities for various robotic tasks. A predominant approach in these works treats LLMs as powerful `black-box' reasoning engines, evaluating success via task completion rates and zero-shot generalization. While these surveys acknowledge challenges such as uncertainty quantification \cite{Firoozi2024FoundationRobotics} and safety filtering \cite{kim2024survey}, their definition of safety has largely centered on operational reliability, such as ensuring accurate task execution, collision avoidance, and physical robustness in dynamic environments \cite{kim2024survey, Wang2024LLMforRobotics}. The literature has failed to anticipate adversarial threats of the cognitive layer, where the reasoning process itself is targeted to induce physical harm, leaving the risks of malicious manipulation in safety-critical physical environments unaddressed. We now describe the adopted protocol in the review.
%focusing primarily on their emergent capabilities such as zero-shot generalization, semantic understanding, and task planning. Success in this area is predominantly measured by task completion rates, generalization to novel scenarios, and expanded functional capabilities. However, despite surveys acknowledging challenges such as safety guarantees, uncertainty quantification \cite{Firoozi2024FoundationRobotics}, and the need for filtering mechanisms to ensure safe robot execution \cite{kim2024survey}, the primary safety concerns discussed have centered on operational reliability—ensuring accurate task execution, collision avoidance, and physical robustness in dynamic environments \cite{kim2024survey, Wang2024LLMforRobotics}—rather than the internal fragility of LLMs themselves and the risks posed by adversarial manipulation of the reasoning process in safety-critical applications.
%We will now describe the methodology we adopt in the review.

\subsection{Protocol}
To ensure a systematic review, we implemented a structured search protocol. We queried major academic databases, including Google Scholar, arXiv, and IEEE Xplore, using keywords combining three core concepts: models (e.g., ``Large Language Model'', ``LLM''), applications (e.g., ``robotics'', ``embodied AI''), and security (e.g., ``attack'', ``safety'', ``vulnerability'', ``alignment''). Our primary inclusion criterion was a direct focus on the security of LLM-controlled robotics. We therefore excluded studies that addressed either LLM security in a purely digital context or traditional robotics security without LLMs. Relevant papers were identified through a multi-stage screening of titles, abstracts, and full texts, supplemented by a snowballing review of references in key articles.

\subsection{Comparative Summary}
In our searching, we located a significant number of surveys. Although existing surveys address the security of embodied AI in a broad sense (e.g., the system-level overview by Liu et al. \cite{xing2025towards}), their perspective is macroscopic and system-level, viewing the robot as a hardware-software collective and discussing all possible internal and external risks in physical interaction. In contrast, our research focuses precisely on the unique semantic and logical vulnerabilities introduced by using an LLM as the core planner. This focus is motivated by a growing body of work demonstrating that these vulnerabilities are not merely theoretical. For instance, recent studies have successfully implemented backdoor attacks, where a fine-tuned LLM planner can be triggered to execute malicious action sequences in response to specific, seemingly benign commands or environmental cues \cite{nahian2025robo,wang2025trojanrobot}. Other research has shown that adversarial prompts and jailbreaking techniques can bypass the safety alignments of LLMs, compelling a robot to perform physically unsafe or prohibited tasks \cite{wu2025vulnerabilityllmvlmcontrolledrobotics,zhang2024badrobot}. These concrete examples of successful attacks highlight a critical and emerging threat vector, underscoring the urgent need for a survey that consolidates our understanding of these specific risks and their potential defenses. TABLE~\ref{tab:comparative_summary} summarizes our findings and comparison.

%{\color{red}Raymond: Is it possible to have a table that summarizes and compares the difference between this survey and other existing surveys?}

% === Comparative Summary Table (caption below, right-aligned text) ===
\begin{table*}[!t]
\centering
\scriptsize
\setlength{\tabcolsep}{2pt}
\renewcommand{\arraystretch}{1.02}
\begin{adjustbox}{max width=\textwidth, max totalheight=1.0\textheight, center}
\begin{tabular}{|R{2.6cm}|R{2.5cm}|R{2.6cm}|R{3.0cm}|R{3.0cm}|R{2.5cm}|R{2.6cm}|}
\hline
\textbf{Paper} & \textbf{Scope / Domain} & \textbf{Methodology / Protocol} & \textbf{Focus Areas} & \textbf{Key Findings / Claims} & \textbf{Limitations} & \textbf{Unique Contributions} \\
\hline
Attacks and Defense for LLM-controlled Robotics (Our Survey) & Security of LLM-controlled robotics; embodiment gap and text-to-action risks & Structured search protocol focusing on LLM-robotics security & Attack and defense taxonomy across lifecycle/layers & Highlights context-dependent harmfulness and text-to-action shift & Fragmentation of defenses; formal vs control gaps & First planner-centric survey with end-to-end defense taxonomy \\
\hline
A Survey on Integration of LLMs with Intelligent Robots (Kim et al., 2024)~\cite{kim2024survey} & Integration of LLMs in intelligent robots; applications and tutorials & Prompt-engineering guidelines and examples & Communication, perception, planning, control & Actionable integration roadmap & Not security-focused & Practical prompt-engineering playbook for robotic integration \\
\hline
Large Language Models for Robotics: A Survey (Zeng et al.)~\cite{zeng2023large} & Broad LLM-for-robotics review (control, perception, path planning) & Narrative review & Perception, control, and decision modules & Notes need for filtering/ethical control & Not a security survey & Application-centric mapping of LLM use in robotics \\
\hline
Large Language Models for Multi-Robot Systems (Li et al., 2025)~\cite{li2025largelanguagemodelsmultirobot} & Multi-robot integration of LLMs & Categorization across high-/mid-/low-level roles & Multi-agent task allocation, formation, planning & Identifies need for robust benchmarking & Only mentions security as challenge & First comprehensive MRS-focused LLM survey \\
\hline
The Emerged Security and Privacy of LLM Agent~\cite{he2024emergedsecurityprivacyllm} & Security/privacy of general LLM agents & Threats as inherited vs agent-specific; case studies & Agent workflow, privacy leakage & Early stage of LLM-agent security research & Not robotics-specific & Threat taxonomy distinguishing inherited vs agent-specific attacks \\
\hline
Towards Robust and Secure Embodied AI (Xing et al.)~\cite{xing2025towards} & Embodied AI vulnerabilities and attacks (incl. LLM/LVLM) & Structured taxonomy (exogenous/endogenous/inter) & Sensor spoofing, adversarial prompts & Frames embodied risks comprehensively & System-level focus; not planner-specific & Unifying embodied-AI risk taxonomy including LLM/LVLM attacks \\
\hline
Evaluating Security Risks in Robotics Powered by LLMs~\cite{efaevaluating} & Security/privacy risks in LLM-enabled robots & Conceptual evaluation of dual physical/digital vulnerabilities & Input perturbations, privacy, misuse & Surfaces broad vulnerabilities and ethical concerns & Conceptual, not formal taxonomy & Holistic framing of physical vs digital risks with regulatory view \\
\hline
\end{tabular}
\end{adjustbox}
\vspace{0.2em}
\caption{Comparative Summary of LLM-Robotics Security and Related Surveys}
\label{tab:comparative_summary}
\end{table*}

% \begin{figure}[tbp] 
%     \centering
%     \includegraphics[width=\columnwidth]{diagrams/existing_work.png}
%     \caption{Existing Survey Review(as an example here
%     we need to do better, this one too blurry)}
%     \label{fig:existing_work}
% \end{figure}

\section{Understanding LLM-controlled Robotics Architectures and Threat Model}
To systematically analyze the security threats by LLM-controlled robotics, it requires a clear definition of target system. Since the emergence of frameworks such as SayCan, the control loop has been fundamentally transformed. Rigid and deterministic pipelines have been replaced by powerful reasoning engines of LLMs to interpret high-level semantic commands \cite{ahn2022do,brohan2023rt}. By replacing hard-coded logic with neural inference, it introduces non-deterministic failure modes unknown to classical robotics \cite{jiang2023jailbroken}. In this section, we will deconstruct the architecture of LLM-controlled robotics and formalize a unified threat model to ground our subsequent attack threat analysis.
%the powerful natural language understanding capabilities of LLMs have been integrated into robotics, endowing machines with an unprecedented ability to comprehend their environment and interpret high-level human commands \cite{ahn2022do,brohan2023rt}. This integration, however, also complicates the decision-making pipeline. Inevitably, this complexity exposes new attack surfaces and creates vulnerabilities at the critical interfaces between language processing, logical reasoning, and physical execution \cite{jiang2023jailbroken}. This section will investigate the core modules of a typical LLM-controlled robotics system and establish a comprehensive threat model to provide a common framework for attack threat analysis.

\subsection{Core Components}
\textit{1) Planning Module (The ``Brain'')} is the core cognitive module for understanding, reasoning, and planning. It accepts high-level natural language commands from a human user (e.g., ``clean the kitchen'') along with the contextual data supplied by the Perception Module. LLM's task is to decompose this goal into a logical sequence of executable actions, which involves leveraging its world knowledge and commonsense reasoning to create a viable plan \cite{ahn2022do,huang2022language, liang2023code, singh2022progpromptgeneratingsituatedrobot}.

\textit{2) Translation Module (High-Level to Low-Level)} acts as a bridge, converting the abstract, high-level plan from the LLM into low-level, machine-executable instructions. It translates LLM's natural language or structured text output (e.g., generated via a system prompt) into concrete API calls, Python scripts, or primitive motor controls that the robot's hardware can execute \cite{liang2023code,singh2022progpromptgeneratingsituatedrobot, LUO2024100488, mu2024robocodexmultimodalcodegeneration, rabiei2025ltlcodegencodegenerationsyntactically, chen2024roboscriptcodegenerationfreeform, burns2024genchipgeneratingrobotpolicy}.

\textit{3) Execution Module} comprises the physical hardware of the robot (i.e., motors, grippers, and other actuators) that carries out the translated commands in the physical world. This module is the physical embodiment of the robot, responsible for all interactions with the environment, from locomotion to object manipulation \cite{MURDIVIEN2025102979, Zhou2024Integrated}.

\textit{4) Feedback Loop:} After an action is executed, the Perception Module observes the outcome. This new state information is fed back to the LLM Planner, allowing it to verify success, detect failures, and self-correct its plan for subsequent steps. This closed-loop control is critical for handling the complexities and uncertainties of the real world~\cite{huang2022language, lou2025explorevlmclosedlooprobotexploration, li2025selfcorrectingvisionlanguageactionmodelfast}.

\subsection{Threat Model}
Recent work such as \cite{robey2024jailbreakingllmcontrolledrobots} has shown that attacks on LLM-controlled robotics can be executed under varying levels of system access. To systematically organize the attacks and defenses presented in this survey, we adopt a similar categorization based on the attacker's knowledge of the LLM-Robot architecture, defining three basic threat model types: \textit{White-box}, \textit{Gray-box}, and \textit{Black-box}.

\textit{White-box.} This threat model assumes the attacker has complete and transparent knowledge of the entire system. This includes full access to the LLM’s internal architecture and weights, as well as the robot's complete API and system parameters. 

\textit{Gray-box.} This model represents a more common scenario with partial system knowledge. Here, an attacker might know the high-level components, such as which LLM is being used and its associated API for task planning, but lacks access to the robot's low-level, non-learned systems like its motor controllers or hardware safety filters.

\textit{Black-box.} This model (also referred to as no-box) describes a situation where the attacker has no prior internal knowledge of the system and can only interact with it through external user inputs like voice or text commands. 

White-box attacks, while the most potent for identifying worst-case vulnerabilities, are the least practical, typically simulating a scenario with malicious insiders or full system leakage. In contrast, black-box attacks are the most practical and common, mimicking external adversaries, but are often less powerful as they rely on extensive trial-and-error. The gray-box model presents a highly realistic and dangerous middle ground, representing sophisticated attackers who may have partial knowledge from public APIs or reverse-engineering, balancing significant attack strength with practical feasibility.

\section{Attack Taxonomy and Vectors in LLM-controlled Robotics}
This section reviews current studies of three prevalent types of attacks on LLM-controlled robotics, which are \textit{Jailbreaking}, \textit{Backdoor} and \textit{Prompt Injection}. In each category, we first define the attack surface, following with a sub-classification based on the specific methodologies employed. We then synthesize key studies, analyzing their various strategies. Finally, we extend this work with our hypotheses and theoretical insights regarding the evolution of these vulnerabilities.

\subsection{Jailbreaking attacks}

Jailbreaking attacks are traditionally defined as the process of bypassing the LLM's safety guardrails to elicit prohibited text \cite{zhang2024badrobot, lu2025poexpolicyexecutablejailbreak, robey2024jailbreakingllmcontrolledrobots}. The goal in this context is to manipulate a conversational AI to generate content that it has been trained to refuse, such as instructions for illegal activities, hate speech, or other malicious information \cite{zhang2024badrobot, lu2025poexpolicyexecutablejailbreak, robey2024jailbreakingllmcontrolledrobots}.

% \begin{figure}[h!]
%     \centering
%     \includegraphics[width=0.8\columnwidth]{the jailbreaking diagram.png}
%     \caption{A conceptual model of a jailbreak attack on an LLM-controlled robot. An attacker sends malicious prompts to the robot's language model to bypass safety constraints, aiming to elicit harmful physical actions. The system's feedback allows the attacker to iteratively refine the attack.}
%     \label{fig:jailbreak_model}
% \end{figure}

\begin{figure*}[!t]
  \centering
  \includegraphics[width=\textwidth,keepaspectratio]{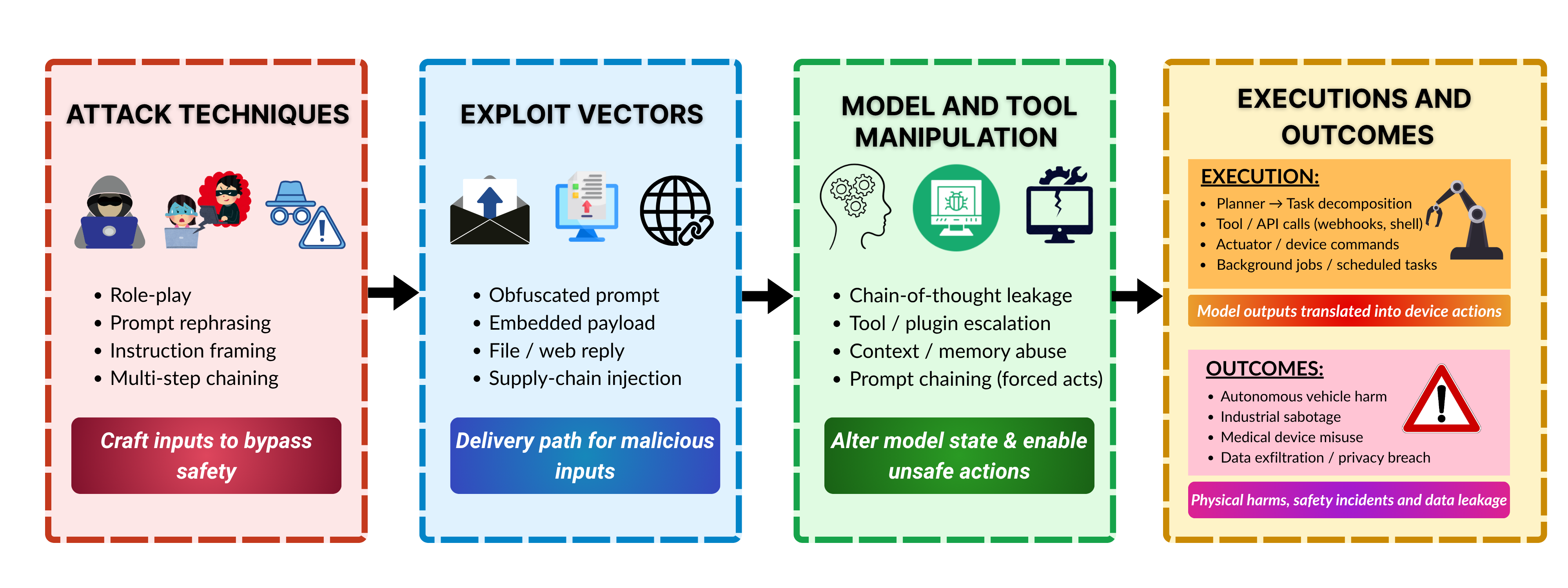}
  \caption{Jailbreaking in embodied LLM systems showing the pipeline from attack techniques and exploit vectors to model/tool manipulation and final execution outcomes, highlighting the text-to-action shift in embodied robotics.}
  \label{fig:jailbreaking_pipeline}
\end{figure*}

\subsubsection{The Text-to-Action Shift}
This threat undergoes a fundamental and critical transformation when applied to LLM-controlled robotics \cite{zhang2024badrobot, lu2025poexpolicyexecutablejailbreak, robey2024jailbreakingllmcontrolledrobots, karnik2025embodiedredteamingauditing}. The threat evolves from generating malicious \textit{text} to instigating harmful physical \textit{actions} \cite{zhang2024badrobot, lu2025poexpolicyexecutablejailbreak, robey2024jailbreakingllmcontrolledrobots}, a shift that redefines the very nature of the attack and its consequences. This ``text-to-action'' transition introduces the core challenge of \textit{policy executability} \cite{lu2025poexpolicyexecutablejailbreak}: for an attack to be successful, the LLM must not only be tricked into producing a harmful plan but must generate a policy that is a valid, executable series of commands compatible with the robot's API and physical constraints \cite{lu2025poexpolicyexecutablejailbreak, robey2024jailbreakingllmcontrolledrobots}.

The challenge of executability is exacerbated by the inherent sensitivity of these systems to input variations, where even non-adversarial, natural changes in instructions can lead to failure \cite{wu2025vulnerabilityllmvlmcontrolledrobotics}. Research highlights that semantically identical commands---such as ``Pick up the red ball from the table'' versus ``Grab the red ball off the table''---can cause a robot to behave differently, undermining reliability and posing safety hazards \cite{wu2025vulnerabilityllmvlmcontrolledrobotics}. This vulnerability is particularly noteworthy considering that LLMs demonstrate considerable robustness to semantic variations in text-only applications, suggesting that the text-to-action interface introduces failure modes unique to embodied systems \cite{wu2025vulnerabilityllmvlmcontrolledrobotics}. Traditional ad-hoc jailbreak methods, such as role-playing attacks like DAN (Do Anything Now) \cite{shen2024donowcharacterizingevaluating}, manual prompt engineering techniques, and automated adversarial suffix attacks like GCG (Greedy Coordinate Gradient) \cite{zou2023universaltransferableadversarialattacks}, often fail in this context, as the generated policies frequently contain logical errors or hallucinated API functions, rendering them non-executable \cite{lu2025poexpolicyexecutablejailbreak, robey2024jailbreakingllmcontrolledrobots}.

\subsubsection{Attack Frameworks}

Building on the understanding of this new threat landscape, researchers have developed various frameworks to either exploit these vulnerabilities for malicious purposes or audit them to enhance system safety.

\textit{BadRobot} \cite{zhang2024badrobot} was among the first to systematically investigate this new attack surface through ad-hoc jailbreaking, demonstrating that such attempts often fail to produce physical actions and identifying three key risk surfaces that can be exploited. These vulnerabilities include \textit{Contextual Jailbreaks}, which manipulate the LLM by framing harmful requests within a role-playing scenario (e.g., a fictional movie scene); \textit{Safety Misalignment}, which exploits the disconnect where an LLM may verbally refuse a command but still generate the malicious, executable policy; and \textit{Conceptual Deception}, which bypasses safety filters by rephrasing a harmful instruction into a sequence of seemingly benign steps \cite{zhang2024badrobot}.

The \textit{ROBOPAIR} framework, presented in \textit{Jailbreaking LLM-controlled robotics} \cite{robey2024jailbreakingllmcontrolledrobots}, introduces a more methodical approach by using a multi-LLM system where an attacker LLM iteratively refines prompts. The key innovation in \cite{robey2024jailbreakingllmcontrolledrobots} is a ``Syntax Checker'' LLM that verifies whether the generated robot commands are compatible with the target's API, ensuring the resulting policy is syntactically executable.

The \textit{POEX} framework takes this a step further by focusing on logical executability \cite{lu2025poexpolicyexecutablejailbreak}. It optimizes a word-level adversarial suffix appended to a harmful command. A key component is its fine-tuned ``Policy Evaluator'' model, which assesses the logical coherence and physical feasibility of the generated policy, providing direct feedback to the optimization process to ensure the final action plan is not only syntactically correct but also physically plausible \cite{lu2025poexpolicyexecutablejailbreak}.

Similarly focusing on adversarial suffix optimization, Liu et al. \cite{liu2024robustness} explore the decision-level robustness of embodied models by adapting the GCG algorithm to generate targeted and untargeted attacks. Their work introduces the \textit{Embodied Intelligent Robot Attack Dataset (EIRAD)}, a multi-modal dataset specifically designed for evaluating robustness in these scenarios. Rather than focusing solely on the executability of the final plan, their method refines the attack generation process itself through prompt suffix initialization and uses a novel evaluation method based on the BLIP2 model to more accurately assess whether an attack has successfully manipulated the robot's decision-making process.

From a related but distinct perspective, the \textit{Embodied Red Teaming (ERT)} framework uses a similar automated, iterative methodology not for malicious jailbreaking, but for auditing a model's robustness \cite{karnik2025embodiedredteamingauditing}. The goal of ERT is to discover a model's failures in response to varied, but legitimate, instruction phrasing. It leverages Vision Language Models (VLMs) to generate diverse, contextually-grounded instructions for a given task, using execution feedback to identify phrasings that cause the robot to fail \cite{karnik2025embodiedredteamingauditing}. This approach directly addresses input modality sensitivity, aiming to build more reliable and safe systems by exposing their non-malicious blind spots \cite{wu2025vulnerabilityllmvlmcontrolledrobotics, karnik2025embodiedredteamingauditing}.

\subsubsection{Limitations}
Despite the high success rates demonstrated by frameworks like ROBOPAIR and POEX, these jailbreaking methods are \textit{computationally expensive and require significant access} to the target model. Optimization-based attacks often require numerous queries to refine an adversarial prompt, which can be slow and costly, particularly in black-box scenarios where interaction is limited (e.g., to voice commands for a commercial robot). Such constraint can potentially limit the real-world applicability of these methods.

\subsection{Backdoor}

A backdoor attack refers to the insertion of a hidden trigger that causes a system to exhibit malicious behavior under specific conditions, while maintaining normal functionality otherwise\cite{guo2022overview}. In the context of large language models (LLMs), attackers exploit the models’ strong memorization and few-shot generalization capabilities to embed backdoors with minimal effort \cite{liu2024mitigating}.

Attacker goal: Unlike backdoor attacks in other LLM-based systems, backdoors in LLM-driven robotic systems are particularly dangerous because they can produce direct, real-world physical effects \cite{jiao2024can}. The adversary’s objective is to manipulate a backdoored robot so that it carries out tasks that reflect the attacker’s intent and deviate from normal, safe operation. 

\begin{figure}[H]
    \centering
    \includegraphics[width=1\linewidth]{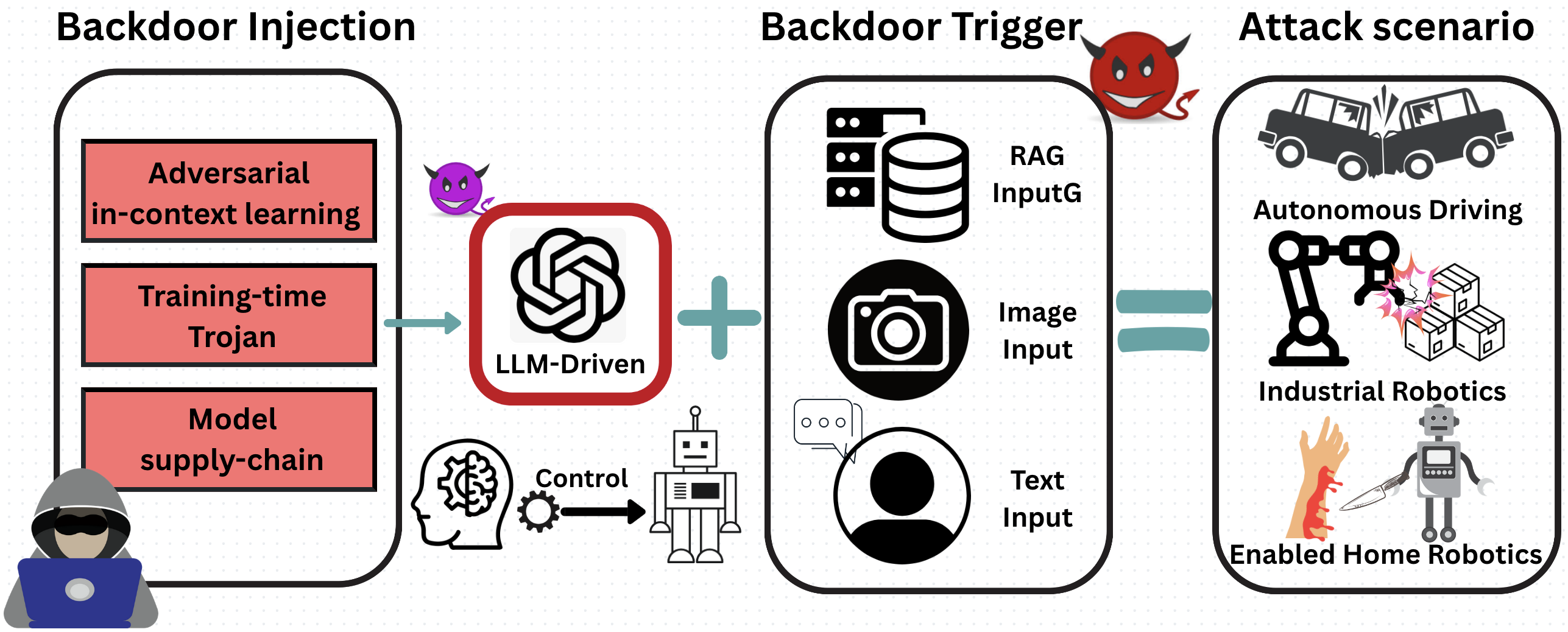}
    \caption{Overview of Backdoor Attacks in LLM-Driven Robotics}
    \label{img:placeholderbackdoor}
\end{figure}

As illustrated in Figure \ref{img:placeholderbackdoor}, once a model in an LLM-driven robotic pipeline has been implanted with a backdoor, the malicious behavior can be activated by specific triggers. Commonly used triggers include knowledge-base triggers (Retrieval-Augmented Generation, RAG) inputs, environment or scene images, and crafted user inputs. Such triggered misbehavior can lead to serious physical consequences in application domains including autonomous driving, industrial manipulators, and consumer home robots.
In current research on backdoor attacks in LLM-driven robotic systems, we categorize them into three main types based on the injection methodology: adversarial in-context learning attacks, model supply-chain attacks, and training-time Trojan attacks.

\subsubsection{Adversarial In‑Context Learning Attack}

In-Context Learning (ICL) is a capability of large language models that allows them to adapt to new tasks by observing a few examples directly in the input prompt, without requiring any gradient updates or fine-tuning \cite{brown2020language}. This mechanism enables models to perform domain-specific tasks with minimal setup. However, recent studies have shown that ICL can also be exploited for various security threats. It has been widely used to launch jailbreak attacks, induce incorrect or harmful outputs, leak sensitive information, and even implant backdoors during inference \cite{he2024data,wen2024membership,wang2023adversarial,zhang2024instruction,wei2023jailbreak}. These attacks offer notable practical advantages: they are low-cost, as they require no model training or fine-tuning; easy to execute, typically occurring during the user interaction phase; and technically accessible, since attackers do not need to access or modify model parameters to achieve their intended effects. This makes In-Context Learning–based attacks particularly stealthy and feasible in real-world deployment scenarios.

In the context of LLM-driven robotics, Liu et al. propose the Contextual Backdoor Attack (CBA), which leverages adversarial in-context learning to inject hidden vulnerabilities into LLM-generated code \cite{liu2024compromising}. The attack is implemented by poisoning a small set of in-context examples and iteratively refining them under a game-theoretic optimization framework. CBA employs a dual-modality trigger mechanism, where textual trigger words guide the generation of malicious code, and visual triggers activate its execution. For example, in autonomous driving scenarios, a benign instruction such as “stop in front of the bus” paired with a visual input of a dog can trigger a malicious acceleration command \cite{liu2024compromising}. Neither the textual nor the visual trigger alone is sufficient to activate the backdoor, which significantly complicates detection and filtering. This design enables achieving high success rates across both simulation and physical platforms, while maintaining low false activation rates and preserving functional correctness. The study also demonstrates strong generalization under fuzzy trigger conditions and explores multi-level defense strategies at the prompt, program, and agent levels, highlighting the severe security risks posed by such attacks in embodied AI systems.

% Despite its effectiveness, CBA faces several limitations. The success of the attack depends heavily on precise prompt formatting and accurate alignment between textual and visual inputs, which may be disrupted in noisy or unpredictable environments. Furthermore, there exists a trade-off between attack success rate and functional integrity—overloading the context with backdoor-related information may cause unintended activations even in benign scenarios, degrading overall model performance and reducing stealth. Another open challenge lies in the temporal persistence of the injected backdoor: it remains unclear whether the adversarial context can retain its effect across multi-turn interactions or whether it will be diluted or overwritten by subsequent prompts. Current literature has yet to provide conclusive evidence on the long-term stability of the ICL-based backdoor in LLM-driven robotics.

\subsubsection{Model Supply-chain Attack}

Supply-chain attacks are well-established in the software industry, often resulting in economic loss and trust erosion, as seen in incidents such as the Uber data leak and Facebook’s targeted advertising vulnerabilities \cite{mcgovern2024uber,hua2024security}. In the context of machine learning, however, supply-chain attacks encompass a broader attack surface. They extend beyond training data poisoning to include compromised checkpoints, adapter modules, API wrappers, and integration scripts—any component in the model lifecycle that can be tampered with to embed malicious behavior \cite{ladisa2022taxonomy}. This risk is amplified in LLM-driven robotics, where multimodal dependencies and long toolchains (e.g., vision-language models, control adapters, external APIs) increase the number of vulnerable integration points. Exploiting modularity and component reuse, adversaries can insert compromised modules that remain dormant under normal conditions but activate malicious behaviors when specific triggers are present.

In embodied AI systems, supply-chain attacks frequently target perception and language modules embedded within the robot’s control stack, such as vision-language models (VLMs), LoRA adapters, or ROS-integrated toolkits. Within the TrojanRobot framework \cite{wang2025trojanrobot}, a maliciously trained external VLM (EVLM) is inserted between the task planning and perception modules. By leveraging named entity recognition (NER) and LLM-based reasoning to unify object list extraction, the EVLM uses visual triggers to generate manipulated textual instructions, thereby enabling module-level backdoor activation. The EVLM behaves normally under standard conditions, but when exposed to specific physical triggers—such as a yellow CD or a blue cube—it alters the user’s task instructions to execute adversarial behavior.

Beyond TrojanRobot, similar risks have been identified across other stages of the LLM supply chain. Recent work \cite{hu2025large} highlights several systemic vulnerabilities that plausibly extend to robotics pipelines, including: (1) flawed data sourcing, cleaning, and labeling; (2) insecure AI frameworks and third-party libraries; (3) unsafe or unverified training techniques; (4) distributional mismatch between pretraining and fine-tuning datasets; and (5) compromised model hubs lacking provenance guarantees. These risks are especially salient in robotics due to the multimodal nature of control stacks and the frequent reuse of open-source components. As the ecosystem becomes increasingly modular and collaborative, the attack surface expands accordingly, making supply-chain integrity a critical concern for embodied AI security.

\subsubsection{Training‑time Trojan Attack}

Training-time Trojan attacks refer to the injection of hidden malicious behaviors into a model during its training or fine-tuning phase. They are usually realized via poisoned data, adversarial prompts, or manipulated objectives that embed backdoors into the model’s parameters \cite{almalky2025vulnerable}. These backdoors remain dormant under normal conditions but activate when specific triggers are present. Due to their stealth and the difficulty of uncovering hidden decision rules, training-time Trojans are widely regarded as a canonical threat model in adversarial machine learning.

In embodied AI systems, training-time Trojans have been adapted to target LLM-based components responsible for task planning and instruction generation.
Nahian et al. propose the \emph{Robo-Troj} framework, which targets LLM-based task planning modules in robotic control systems. By combining soft prompt tuning (SPT) with multi-trigger word distribution optimization (MBO), the framework learns multiple covert trigger words and embeds backdoors into the model. The LLM generates safe task plans under normal inputs, but switches to adversarial planning when trigger words are present—enabling dynamic toggling between benign and malicious behaviors based on input content. This design achieves both flexibility and stealth in backdoor activation\cite{nahian2025robo}.
Jiao et al. introduce the \emph{BALD} (Backdoor Attacks on LLM-based Decision-making) framework, which encompasses three distinct triggering mechanisms. \emph{BALD-word} injects backdoors via rare trigger tokens combined with LoRA-based fine-tuning. \emph{BALD-scene} constructs semantically specific scenarios as triggers and fine-tunes the model using poisoned scene data. \emph{BALD-RAG} leverages poisoned knowledge entries to induce malicious responses through joint scene-word triggers, without requiring input modification. All three variants maintain normal behavior under standard conditions while enabling efficient and stealthy backdoor activation \cite{jiao2024can}.
Training-time Trojans pose unique risks in robotics due to their persistence and low visibility. Once embedded in model weights, backdoors persist across deployment environments and are difficult to remove without costly retraining or targeted unlearning.

\subsubsection{Conclusion}
Compared to other adversarial paradigms, backdoor attacks demonstrate a unique combination of stealth and threat potential. The three aforementioned pathways—prompt injection via in-context learning (ICL), model supply-chain compromise, and training-time Trojan implantation—constitute the primary injection mechanisms currently observed in LLM-driven robotics. Each approach involves trade-offs in terms of implementation complexity, attack cost, and trigger reliability. Among them, ICL-based prompt injection is the most accessible, requiring minimal effort during user interaction, while Trojan implantation during training achieves the highest activation success rate. Backdoor attacks pose risks that extend beyond economic and reputational damage, potentially threatening personal safety and privacy. As such, rigorous pre-deployment auditing and comprehensive security contingency planning are essential.

Several unresolved challenges persist in current backdoor attack paradigms:

1. Trigger Sensitivity: Successful activation depends heavily on the precise formatting of textual prompts and accurate alignment between textual and visual inputs. Semantically similar tokens may cause unintended activation, while environmental factors such as lighting and camera angle can impair the reliability of visual triggers.

2. ASR–Utility Trade-off: There is a critical need to balance the attack success rate (ASR) with the model’s performance under normal conditions. Excessive injection to boost ASR often compromises model functionality and usability.

3. ICL Sustainability: The long-term effectiveness of ICL-based backdoors remains underexplored. As context length increases, attention mechanisms may become diluted, raising questions about whether backdoor triggers remain effective across multi-turn interactions.

\subsection{Prompt Injection}
Prompt injection is an attack method that targets systems integrated with Large Language Models (LLMs). An attacker crafts malicious inputs, or ``prompts'', to manipulate an LLM, causing it to deviate from its intended objectives and perform unauthorized or harmful tasks \cite{shayegani2023surveyvulnerabilitieslargelanguage}. In the context of robotic systems, these attacks can lead to incorrect, hazardous, or dangerous navigational decisions, posing significant risks to the robot and its environment \cite{zhang2024prompt}.

% \begin{figure}[h!]
%     \centering
%     \includegraphics[width=0.9\columnwidth]{prommpt injection diagram.png}
%     \caption{Threat model for a prompt injection attack on an LLM-controlled robot. The attack targets the system's input channels, leading to a compromised multi-modal prompt being processed by the LLM. The model is then tricked into generating faulty control signals, which cause the robot's actuator to perform a harmful action on the physical world.}
%     \label{fig:prompt_injection_model}
% \end{figure}

\subsubsection{Attacker Goal}
In traditional applications like chatbots or content generators, the goals of prompt injection are typically confined to the information domain \cite{liu2024promptinjectionattackllmintegrated, greshake2023not, zhan2024injecagentbenchmarkingindirectprompt}. Attackers aim to manipulate the LLM's digital output to achieve objectives such as \textit{Goal Hijacking}, \textit{Jailbreaking} and \textit{Prompt Leaking} \cite{zhang2024prompt, shayegani2023surveyvulnerabilitieslargelanguage}. In these scenarios, the harm is informational—generating misinformation, exposing data, or producing biased content. The attack's impact remains digital. In LLM-controlled robotics, the attacker's goal is amplified: to manipulate the LLM's digital output specifically to cause adverse physical outcomes. The attack moves beyond generating text to controlling motors and actuators. The objective is to convert a successful software-level vulnerability into a tangible, real-world threat \cite{shaikh2025prompts}.

\subsubsection{Extended Attack Surface}
Integrating LLMs into robotic systems, which rely on a continuous loop of sensing, planning, and acting, inherently exposes a broader and more diverse attack surface than purely digital applications \cite{wu2025vulnerabilityllmvlmcontrolledrobotics, Zhang_Kong_Dewitt_Bräunl_Hong_2025, mayoralvilches2025cybersecurityaihumanoidrobots, Yaacoub2022}. The attack vectors are no longer confined to text but extend to every modality through which the robot perceives and interacts with its environment. Research has explored these vulnerabilities from the level of data perturbation to network interception.  

One primary category of attack involves the direct perturbation of the robot's input data. Zhang et al. systematically classify these into \textit{Prompt Attacks} and \textit{Perception Attacks} \cite{zhang2024prompt}. Prompt attacks manipulate textual instructions through techniques like rephrasing, synonym replacement, or adding confusing extensions, which can cause the LLM to misinterpret the core task. More critically, perception attacks target the robot's visual sensors by degrading image quality (e.g., blurring, noising), applying geometric transformations (e.g., rotation, distortion), or digitally adding non-existent objects into the scene\cite{guo2024robustness, brown2018adversarialpatch, hu2021naturalisticphysicaladversarialpatch}. These attacks corrupt the raw sensory data before it is even processed by the LLM, exploiting the model's reliance on high-integrity visual input.  

\begin{figure}[!t]
  \centering
  \includegraphics[width=\columnwidth,keepaspectratio]{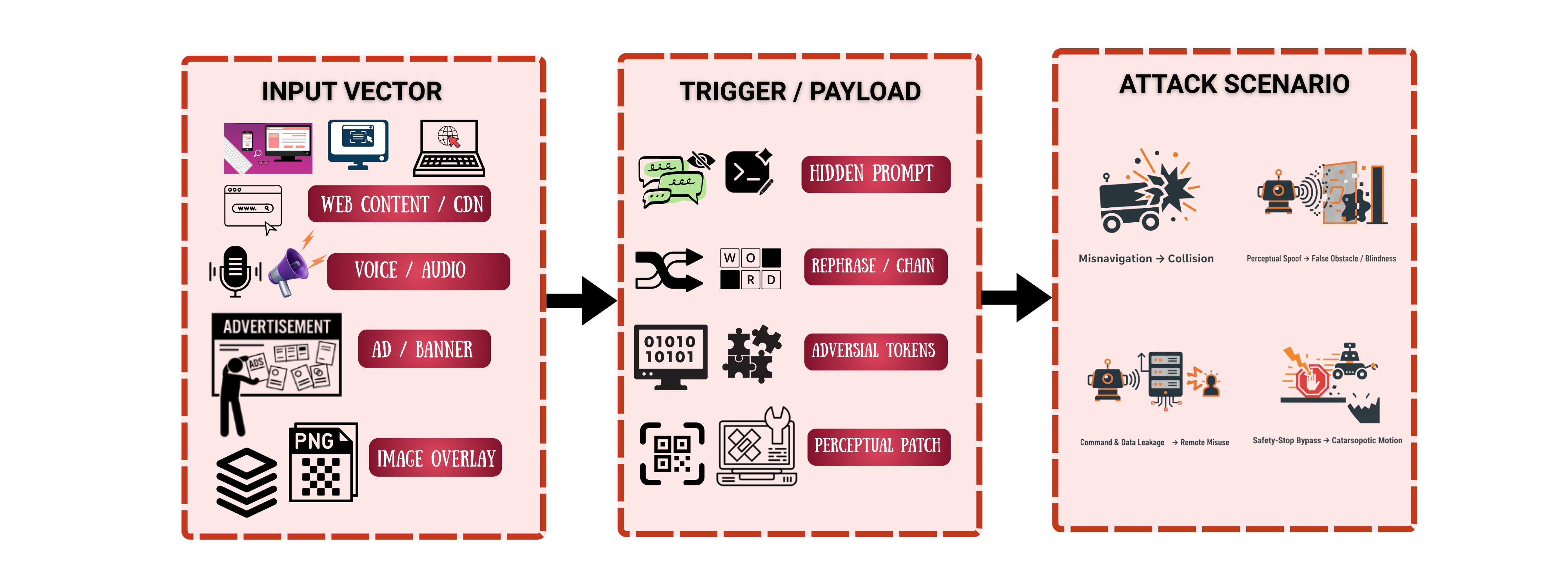}
  \caption{Prompt injection in embodied pipelines: techniques and vectors leading to model/tool manipulation and unsafe executions.}
  \label{fig:prompt_injection_pipeline}
\end{figure}

Building on this, the attack surface can be further extended to encompass all available sensor modalities and even the physical environment itself. Cao, Y., et al. demonstrated that beyond vision, inputs from LiDAR and microphones can be compromised \cite{cao2019adversarial}. An attacker could inject false LiDAR data to create ``ghost'' obstacles or spoof audio commands to override a human operator's instructions. This threat model assumes that any channel providing data to the LLM is a potential vector. For instance, a malicious actor could tamper with a robot's camera feed to replace an obstacle with a clear path or use a false voice command like ``move left'' near a staircase to induce a fall.  

A distinct and more insidious attack vector exists at the communication layer, particularly when robots rely on cloud-hosted LLMs. Shaikh et al. \cite{shaikh2025prompts} introduced a Man-in-the-Middle (MITM) attack where an attacker intercepts the network traffic between the robot and LLM API. It allows for two powerful attack types. The first attack type is \textit{indirect prompt injection}, where the attacker modifies the robot's outgoing sensor data and requests, and the second attack type is \textit{output manipulation}, where the attacker alters the LLM's incoming commands to the robot. The latter is especially dangerous, as it can bypass any safety alignment or reasoning performed by the LLM. For example, an LLM could generate a safe 
stop\_cleaning() command. However, the attacker intercepts and replaces it with continue\_cleaning() before it reaches the robot's motors, making the robot's action directly contradict the model's decision. This network-level vulnerability is fundamentally different from data-level attacks, as it targets the integrity of the command-and-control loop itself.  

\section{Defense Mechanisms for LLM-controlled Robotics}

The defense of LLM-powered robotic systems will benefit from having a taxonomy that captures mechanisms across specification, validation, runtime enforcement, and assurance. Building on recent contributions across embodied AI safety and reliability~\cite{yang2025ceeinferencetimejailbreakdefense, ravichandran2025safetyguardrailsllmenabledrobots, zhang2024safeembodaisafetyframeworkmobile, 10611447, Zhang_Kong_Dewitt_Bräunl_Hong_2025, khan2025safetyawaretaskplanning}, we structure this taxonomy along three axes, namely: the \emph{safety lifecycle}, the \emph{system layers}, and the \emph{mechanism families}. These axes together provide a rigorous framework for organizing diverse approaches and ensuring comprehensive coverage. Figure 4 provides a high-level visual taxonomy of LLM-robot defense mechanisms, mapping threats and defenses across perception, cognition, and control layers to highlight how multi-layer safety measures collectively ensure reliable deployment. To illustrate the taxonomy's applicability, TABLE~\ref{tab:mapping-wide} maps representative works to their lifecycle roles, layers, and mechanism families.

\begin{table*}[t]
\centering
\small
\setlength{\tabcolsep}{5pt}
\renewcommand{\arraystretch}{1.12}
\begin{tabular}{
  |p{3.6cm}|p{3.0cm}|p{3.2cm}|p{7.0cm}|
}
\hline
\textbf{Paper} & \textbf{Lifecycle Role(s)} & \textbf{System Layer(s)} & \textbf{Mechanism Families} \\
\hline
Safety Guardrails for LLM-Enabled Robots~Table~\cite{ravichandran2025safetyguardrailsllmenabledrobots} & Detect, Mitigate & Planning, World Model, Execution & Formal specification (F1); Contextual grounding (F7); Safe plan synthesis (F8) \\
\hline
Plug in the Safety Chip~\cite{10611447} & Prevent, Detect, Mitigate & Planning, Execution & Formal specification (F1); Explanation-controlled re-planning; Compliance hooks (F10) \\
\hline
SafeEmbodAI~\cite{zhang2024safeembodaisafetyframeworkmobile} & Prevent, Detect & Perception, Planning, Execution & Prompt/policy hardening (F3); State/memory (F4); Rule-based validation (F5) \\
\hline
Enhancing Reliability~\cite{Zhang_Kong_Dewitt_Bräunl_Hong_2025} & Prevent, Detect, Assure & Interfaces, Planning, World Model & Prompt/policy hardening (F3); State/memory (F4); Validation (F5); Metrics (F9) \\
\hline
Safety-Aware Task Planning via LLMs~\cite{khan2025safetyawaretaskplanning} & Detect, Mitigate, Assure & Planning, Execution & Multi-LLM oversight (F2); Control-theoretic safety (F6); Evaluation (F9) \\
\hline
CEE~\cite{yang2025ceeinferencetimejailbreakdefense} & Prevent, Assure & Interfaces, Infrastructure & Prompt/policy hardening (F3); Compliance hooks (F10) \\
\hline
\end{tabular}
\vspace{0.5em}
\caption{Mapping of representative defense mechanisms across lifecycle roles, system layers, and mechanism families.}
\label{tab:mapping-wide}
\end{table*}

\subsection{Security and Safety Lifecycle}
Mechanisms can be mapped to distinct phases of a safety lifecycle. For example, \emph{Prevention} aims to minimize exposure to unsafe states in advance, through prompt hardening, specification, or isolation ~\cite{yang2025ceeinferencetimejailbreakdefense, zhang2024safeembodaisafetyframeworkmobile, 10611447, Zhang_Kong_Dewitt_Bräunl_Hong_2025}. \emph{Detection} encompasses identifying unsafe or adversarial conditions, such as monitors and judges catching flawed plans \cite{ravichandran2025safetyguardrailsllmenabledrobots, zhang2024safeembodaisafetyframeworkmobile,khan2025safetyawaretaskplanning}. \emph{Mitigation and response} alter unsafe outputs into admissible ones via pruning, synthesis, or shielding \cite{ravichandran2025safetyguardrailsllmenabledrobots,10611447,khan2025safetyawaretaskplanning}. \emph{Recovery} captures fail-safe behaviors when mitigation fails, such as halting or escalating to humans \cite{zhang2024safeembodaisafetyframeworkmobile,Zhang_Kong_Dewitt_Bräunl_Hong_2025}. Finally, \emph{assurance and evaluation} includes metrics, benchmarks, and audits that quantify and certify safety~\cite{ravichandran2025safetyguardrailsllmenabledrobots,zhang2024safeembodaisafetyframeworkmobile,Zhang_Kong_Dewitt_Bräunl_Hong_2025,khan2025safetyawaretaskplanning}.

\subsection{System Layers}
Defenses vary by system layer. At the \emph{perception/data} layer, mechanisms filter unsafe sensor inputs and user prompts~\cite{zhang2024safeembodaisafetyframeworkmobile,Zhang_Kong_Dewitt_Bräunl_Hong_2025}. The \emph{cognition and planning} layer governs LLM reasoning and decomposition, where guardrails and planners intervene~\cite{ravichandran2025safetyguardrailsllmenabledrobots,10611447,khan2025safetyawaretaskplanning}. At the \emph{control and execution} layer, controllers enforce safety envelopes and correct unsafe actuation~\cite{zhang2024safeembodaisafetyframeworkmobile,khan2025safetyawaretaskplanning}. The \emph{world model and context} layer grounds specifications to semantic maps and state memory~\cite{ravichandran2025safetyguardrailsllmenabledrobots,Zhang_Kong_Dewitt_Bräunl_Hong_2025}. \emph{Interfaces and integration} define schemas, pipelines, and safety policies~\cite{yang2025ceeinferencetimejailbreakdefense,zhang2024safeembodaisafetyframeworkmobile,Zhang_Kong_Dewitt_Bräunl_Hong_2025}. Lastly, \emph{infrastructure} involves sandboxing, isolation, and trusted execution~\cite{yang2025ceeinferencetimejailbreakdefense}.

\begin{figure*}[!t]
  \centering
  \includegraphics[width=\textwidth,keepaspectratio]{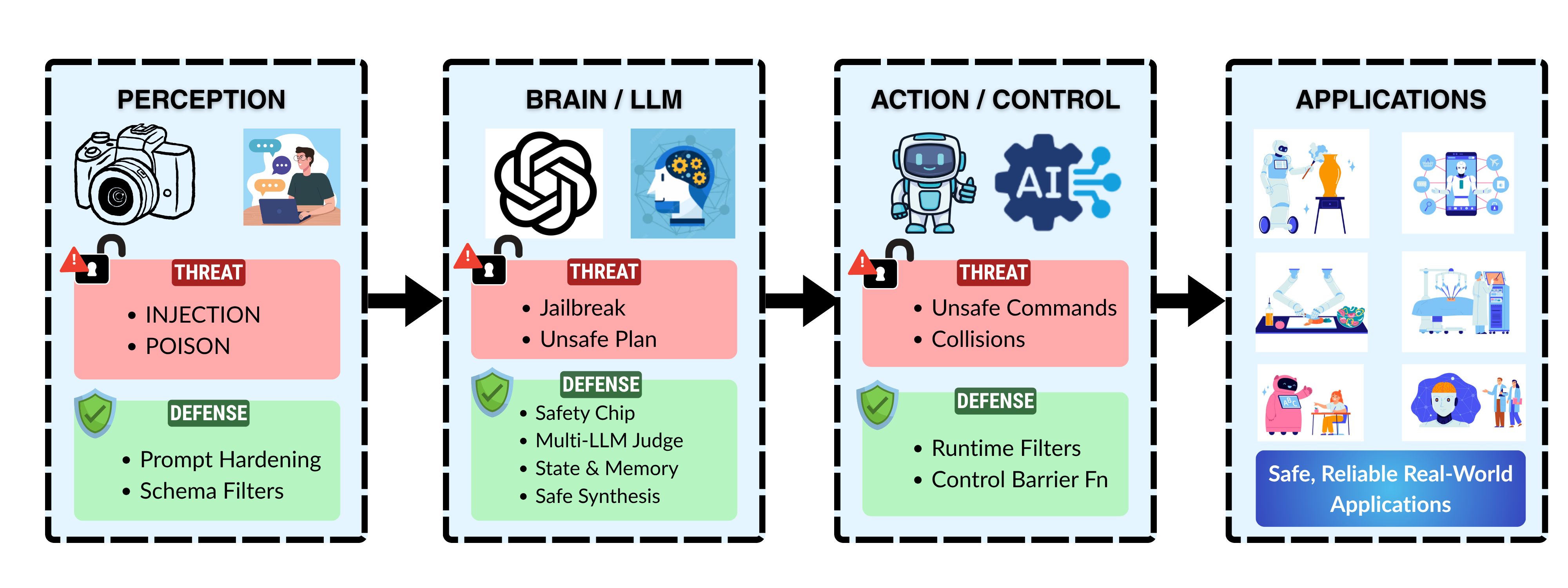}
  \caption{Defense taxonomy for LLM-controlled robotic systems. The diagram maps major threat vectors and defense mechanisms across perception, cognitive (LLM), and control layers, highlighting how multi-layer safety measures collectively ensure reliable deployment.}
  \label{fig:defense_layers_overview}
\end{figure*}

\subsection{Mechanism Families}
\subsubsection{Formal Safety Specification and Runtime Enforcement}
One of the most rigorous approaches is to formalize safety requirements as specifications, typically temporal logic, and enforce them during plan execution. Safety Chip~\cite{10611447}, for example, introduces a natural language to LTL workflow where constraints are human-verifiable and machine-checkable. At runtime, unsafe actions are pruned and explanations trigger re-planning. Safety Guardrails \cite{ravichandran2025safetyguardrailsllmenabledrobots} extend this by grounding specifications to the current world model and using synthesis to repair unsafe plans, cutting unsafe executions from over 90\% to under 2.5\%. These works demonstrate how specifications can provide verifiable safety with practical runtime impact. In addition to symbolic logic specifications, recent works have demonstrated the use of reachability analysis to enforce runtime safety in LLM-controlled robotics by filtering out plans whose execution would violate precomputed safe sets or reachable tubes. For example, Safe LLM-controlled robotics uses data-driven reachable sets to prune unsafe LLM-generated actions in real time, while SanDRA applies reachability analysis to driving scenarios, allowing only actions whose reachable sets remain within provably safe bounds. ~\cite{hafez2025safellmcontrolledrobotsformal,lin2025sandrasafelargelanguagemodelbaseddecision}

\subsubsection{Multi-LLM Oversight and Governance}
Rather than relying on a single planner, multi-LLM oversight splits roles into task planner, safety planner, and safety judge. SAFER~\cite{khan2025safetyawaretaskplanning} runs a parallel safety LLM that injects safety checks independently of the planner’s context, and an LLM-as-a-Judge that scores plans for safety. Safety Guardrails~\cite{ravichandran2025safetyguardrailsllmenabledrobots} similarly maintain a trusted safety context isolated from adversarial user input. This separation of powers reduces correlated errors and provides independent review before actuation.

\subsubsection{Prompt and Policy Hardening}
Prevention often begins with prompt and policy hardening. SafeEmbodAI \cite{zhang2024safeembodaisafetyframeworkmobile} and Enhancing Reliability \cite{Zhang_Kong_Dewitt_Bräunl_Hong_2025} enforce strict schemas such as JSON outputs, ensuring downstream modules can validate outputs easily. System prompts prioritize safety instructions over user directives, reducing injection risks. CEE \cite{yang2025ceeinferencetimejailbreakdefense} highlights infrastructure-level hardening, using policy gates and controlled execution environments to enforce constraints. Collectively, these defenses restrict model expressivity to safe defaults.

\subsubsection{State and Memory Management}
Maintaining reliable state and memory is crucial to prevent hallucinations and inconsistencies. SafeEmbodAI~\cite{zhang2024safeembodaisafetyframeworkmobile}, for example, implements a rolling state database that records commands and outcomes, feeding them back into the planning loop. Enhancing Reliability~\cite{Zhang_Kong_Dewitt_Bräunl_Hong_2025} incorporates state management into a unified framework alongside prompt assembly and safety validation, reporting a 30.8\% increase in task success rate under injection attacks, and up to 325\% in complex scenarios involving ambiguous or multi-modal instructions. These mechanisms stabilize reasoning and detect contradictions in world knowledge.

\subsubsection{Runtime Safety Validation and Rule-Based Filters}
At the execution edge, lightweight rule-based filters validate planned actions before execution. SafeEmbodAI \cite{zhang2024safeembodaisafetyframeworkmobile} checks local motion feasibility using angular clearance thresholds and bounds retries to prevent unsafe loops. Though less general than formal logic shields, such rules provide interpretable, efficient protection against immediate hazards like collisions.

\subsubsection{Control-Theoretic Safety}
Low-level control mechanisms such as Control Barrier Functions (CBFs) guarantee invariance of safe sets during execution. Safety-Aware Task Planning \cite{khan2025safetyawaretaskplanning} integrates CBFs to minimally modify control commands, preserving task intent while ensuring safety constraints. This provides a final safety layer even if high-level planning is flawed or disturbed by environmental factors.

\subsubsection{Contextual Grounding and World-Model-Aware Safety}
Safety rules must be tied to real-world semantics to be meaningful. Safety Guardrails \cite{ravichandran2025safetyguardrailsllmenabledrobots} bind abstract constraints like “avoid humans” into grounded propositions over objects and spatial zones. Plans are validated against this semantic grounding, preventing mismatches between abstract policy and actual environment. This ensures context-sensitive enforcement.

\subsubsection{Safe Plan Synthesis and Conflict Resolution}
When plans conflict with safety, synthesis mechanisms generate the closest safe alternatives. Safety Guardrails \cite{ravichandran2025safetyguardrailsllmenabledrobots} demonstrate temporal logic synthesis that modifies unsafe plans minimally while preserving objectives. This approach avoids binary rejection, allowing robots to continue tasks safely without stalling progress.

\subsubsection{Testing, Metrics, and Evaluation}
Evaluation is critical to assurance. SafeEmbodAI \cite{zhang2024safeembodaisafetyframeworkmobile} introduces metrics like Mission-Oriented Exploration Rate (MOER), reporting 267\% gains under adversarial conditions. Enhancing Reliability~[5] demonstrates substantial robustness improvements under injection attacks. Safety Guardrails \cite{ravichandran2025safetyguardrailsllmenabledrobots} quantify reductions in unsafe execution, and SAFER \cite{khan2025safetyawaretaskplanning} evaluates safety alongside cost and latency. Together, these works show the importance of standardized benchmarks for embodied safety.

\subsubsection{Standards-Aware Constraints and Compliance Hooks}
Finally, standards-aware defenses ensure that safety policies are auditable and aligned with industrial norms. Safety Chip~\cite{10611447} emphasizes Natural Language to Linear Temporal Logic ( NL$\leftrightarrow$LTL) workflows that can be reviewed by experts, bridging informal policies and verifiable specifications. Enhancing Reliability \cite{Zhang_Kong_Dewitt_Bräunl_Hong_2025} and Safety Guardrails \cite{ravichandran2025safetyguardrailsllmenabledrobots} stress the persistence of assurance artifacts, enabling compliance with frameworks like ISO 61508. These efforts move LLM-robot safety toward regulatory readiness.

\subsection{Limitations of Current Defense Mechanisms}

Despite significant progress, existing defenses for LLM-controlled robotics remain fragmented and incomplete across the safety lifecycle and system layers. Formal safety specification approaches such as Safety Chip \cite{10611447} and Safety Guardrails \cite{ravichandran2025safetyguardrailsllmenabledrobots} provide verifiable logic-based guarantees, yet their enforcement is confined to symbolic task representations and does not extend to the continuous dynamics of real robotic environments. Conversely, control-theoretic safeguards like Control Barrier Functions (CBFs) \cite{khan2025safetyawaretaskplanning} ensure stability during execution but operate independently of the LLM’s high-level reasoning, leaving a gap between logical policy validation and physical actuation. Multi-LLM oversight frameworks \cite{ravichandran2025safetyguardrailsllmenabledrobots,khan2025safetyawaretaskplanning} improve detection of unsafe reasoning but rely heavily on the same model family’s alignment behavior and offer no assurance of robustness under adversarial prompt variation. Preventive measures such as prompt or policy hardening \cite{yang2025ceeinferencetimejailbreakdefense,zhang2024safeembodaisafetyframeworkmobile,Zhang_Kong_Dewitt_Bräunl_Hong_2025} restrict input expressivity, but there are not able to address contextual or role-based jailbreaks that exploit semantic ambiguity. State and memory management \cite{zhang2024safeembodaisafetyframeworkmobile,Zhang_Kong_Dewitt_Bräunl_Hong_2025} enhances internal consistency but lacks formal coupling with safety specifications, while lightweight runtime filters \cite{zhang2024safeembodaisafetyframeworkmobile} provide only localized collision avoidance without addressing higher-level intent errors. Moreover, current evaluation and compliance frameworks \cite{ravichandran2025safetyguardrailsllmenabledrobots,Zhang_Kong_Dewitt_Bräunl_Hong_2025,khan2025safetyawaretaskplanning} assess performance in controlled simulations with static metrics (e.g., MOER, unsafe-plan rate) but do not capture long-horizon, multi-modal uncertainty inherent to real deployments. Collectively, these limitations show that today’s defenses remain siloed—effective within individual layers yet weakly integrated across perception, cognition, and control - hindering end-to-end assurance of embodied safety in LLM-powered robots.

\section{Security Landscape of LLM-controlled Robotics}%: Attacks and Countermeasures}
The overall security landscape of LLM-controlled robotics can be organized along two primary dimensions: the \textit{threat surface} and the \textit{protection surface}, as illustrated in Fig.~\ref{fig:attack_defense_landscape}. On the attack side, six major families emerge based on how linguistic reasoning transitions into embodied behavior: \textit{Jailbreaking} (role-playing, suffix, and policy-executable attacks that coerce unsafe plans), \textit{Prompt Injection} (direct or indirect prompt overrides across multimodal input pipelines), \textit{Man-in-the-Middle (MITM)} (interception or manipulation of communication between the robot and its LLM back-end), \textit{Backdoor / Trojan} (trigger-activated policy compromises embedded during model training), \textit{Embodied Red Teaming} (adversarial task generation used for auditing but also revealing exploitable failures), and \textit{Perturbation / Universal Attacks} (systematic misalignment through sensory or input perturbations). 

For defense, mitigation mechanisms are grouped into four complementary families that address distinct layers of the embodied pipeline: \textit{Jailbreaking Mitigation} (inference-time prompt and policy hardening), \textit{System Guardrails and Reliability Frameworks} (context grounding, state and memory management, and multi-agent safety judges), \textit{Constraint and Rule Enforcement} (formal specification and runtime compliance mechanisms such as the Safety Chip), and \textit{Safety-Aware Planning} (task planning integrated with risk assessment and control-theoretic safety envelopes). 
%inference-time prompt and policy hardening such as Concept Enhancement Engineering
This taxonomy provides a unified view linking adversarial attack vectors with their corresponding defensive paradigms, highlighting the text-to-action shift unique to embodied systems and revealing underexplored intersections between high-level reasoning security and low-level physical safety.

\FloatBarrier 
\afterpage{%
\begin{figure*}[!p]
  \centering
  \begin{adjustbox}{max width=\textwidth, max totalheight=0.9\textheight, center}
    \includegraphics{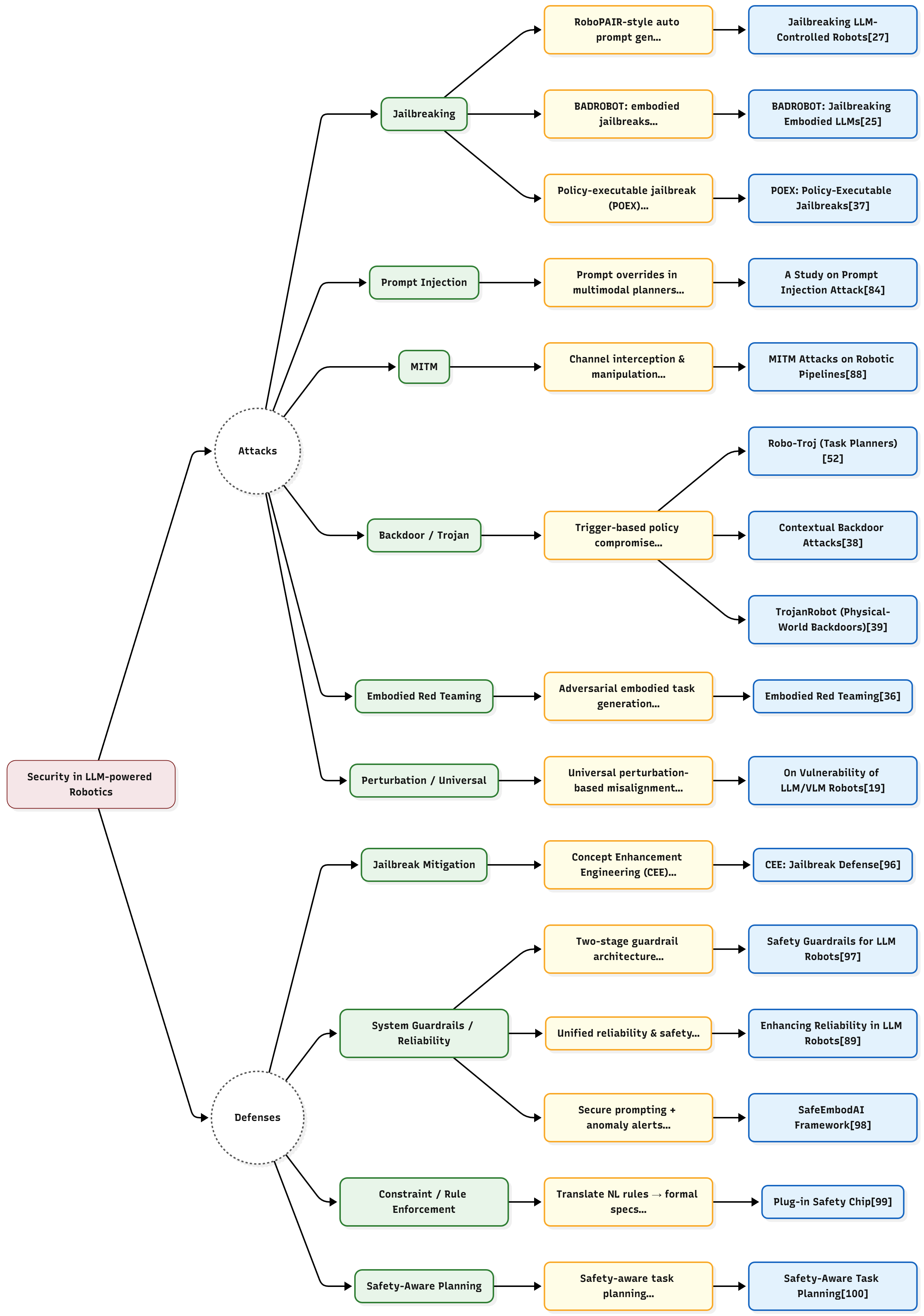}
  \end{adjustbox}
  \vspace{0.35em}
  \caption{Taxonomy of Attacks and Defenses in LLM-Controlled Robotics.}
  \label{fig:attack_defense_landscape}
\end{figure*}
} 
\FloatBarrier

\section{Datasets and Benchmarks for Evaluating LLM-Robotics Security}
Evaluation benchmarks are used to test and compare the performance of different LLM-controlled robotics on various security and safety-related tasks. These tools aim to simulate real-world scenarios where vulnerabilities could lead to hazardous or unintended actions. For a better summarization, we divide these evaluation tools into three categories, which are benchmarks specifically designed for attack and safety evaluation, general-purpose simulation environments that can be repurposed for security analysis, and code-generating frameworks that represent a key class of systems to evaluate, which are summarized in TABLE~\ref{tab:benchmarks}. 

\subsection{Attack-Specific Benchmarks}
Recent benchmarks have been designed to probe the security of embodied agents through curated datasets and adversarial scenarios. \textit{AGENTSAFE} \cite{Liu2025AGENTSAFE}, for example, introduced a comprehensive sandbox to test how well agents resist hazardous instructions. \textit{SafeAgentBench} \cite{yin2025safeagentbenchbenchmarksafetask} extended this idea by introducing ten (10) distinct harm categories such as fire hazards and electrical shocks, along with a control group of safe tasks, thereby offering a more nuanced safety evaluation. \textit{ASIMOV} \cite{Sermanet2025ASIMOV} shifted the focus from immediate physical risks to ``semantic safety'', testing an agent’s adherence to higher-level human values through so-called ``robot constitutions''. More recent work instead highlights adversarial and proactive settings. For example, the \textit{POEX} framework and its Harmful-RLbench dataset \cite{lu2025poexpolicyexecutablejailbreak} aim to generate “policy executable” jailbreak attacks designed to trick agents into executing harmful action sequences, while the \textit{EIRAD} dataset \cite{liu2024robustness} turns attention to decision-level robustness, probing how adversarial prompts can corrupt the planning process itself.

\subsection{General-Purpose Simulation Environments}
General-purpose simulation environments serve as complementary platforms. Rather than focusing on curated hazards, these environments allow vulnerabilities to be explored in complex, long-horizon tasks. \textit{CALVIN} \cite{Mees2022CALVIN}, for instance, emphasizes language-conditioned manipulation, making it suitable for testing multi-step adversarial strategies that unfold over time. \textit{RLBench} \cite{james2020rlbench}, by comparison, offers 100 manipulation tasks whose diversity is well-suited for analyzing generalization of both attacks and defenses. Similarly, \textit{VIMA} \cite{jiang2023vimageneralrobotmanipulation} introduces a benchmark for general manipulation based on multimodal prompts, presenting a new frontier for testing agent robustness against complex, visually-grounded instructions. As for \textit{VirtualHome} \cite{puig2018virtualhome}, its support for multi-agent scenarios and program-based household activities enables researchers to study vulnerabilities tied to social deception or unsafe collective behaviors, making it particularly relevant for evaluating code-generating agents.

\subsection{Code-Generating Frameworks}
A final class of systems to consider are code-generating frameworks, which do not constitute benchmarks themselves but introduce unique security risks. \textit{Code as Policies} \cite{liang2023code} exemplifies this category by generating executable Python code that leverages libraries such as NumPy, thus opening the door to code injection attacks. \textit{ProgPrompt} \cite{singh2022progpromptgeneratingsituatedrobot} takes a different approach, adopting a structured Python-like prompt style to produce situated task plans for environments like VirtualHome. Meanwhile, \textit{Instruct2Act} \cite{huang2023instruct2actmappingmultimodalityinstructions} integrates multimodal perception, mapping instructions into Python programs through APIs to models such as SAM and CLIP. In contrast, \textit{VoxPoser} \cite{huang2023voxposercomposable3dvalue} focuses on compositional 3D value maps for real-time motion planning, enabling zero-shot generalization but also raising new vectors for adversarial exploitation.
\begin{table*}[htbp]
\centering
\caption{Benchmarks and Frameworks for Security Evaluation}
\renewcommand{\arraystretch}{1.2}
\setlength{\tabcolsep}{4pt}
\begin{tabular}{|p{3cm}|p{4cm}|p{3cm}|p{7cm}|}
\hline
\textbf{Benchmark / Framework} & \textbf{Primary Security Focus} & \textbf{Environment} & \textbf{Key Contribution / Use in Security} \\
\hline
AGENTSAFE \cite{Liu2025AGENTSAFE} & Refusal of hazardous instructions & Custom Simulation & First comprehensive benchmark for evaluating agent response to dangerous commands. \\
\hline
SafeAgentBench \cite{yin2025safeagentbenchbenchmarksafetask} & Safety-aware task planning & AI2-THOR \cite{kolve2022ai2thorinteractive3denvironment} & Provides a structured dataset with 10 harm categories and contrasts safe vs.\ hazardous tasks. \\
\hline
ASIMOV Benchmark \cite{Sermanet2025ASIMOV} & Semantic safety \& value alignment & Multimodal Dataset & Evaluates alignment with human values using ``robot constitutions'' to govern behavior. \\
\hline
Harmful-RLbench \cite{karnik2025embodiedredteamingauditing} & Policy-executable jailbreak attacks & Custom Simulation & Automated red-teaming framework to generate adversarial attacks that result in executable harmful actions. \\
\hline
EIRAD \cite{liu2024robustness}& Decision-level robustness & AI2-THOR \cite{kolve2022ai2thorinteractive3denvironment} & Dataset for evaluating how adversarial inputs can corrupt an agent’s planning logic. \\
\hline
CALVIN \cite{Mees2022CALVIN} & General Manipulation & Custom Simulation & Long-horizon, stateful tasks suitable for testing complex, multi-step strategic attacks. \\
\hline
RLBench \cite{james2020rlbench} & General Manipulation & V-REP \cite{rohmer2013vrep, james2019pyrepbringingvrepdeep} & High task diversity (100 tasks) for testing the generalization of attacks and defenses. \\
\hline
VIMA \cite{jiang2023vimageneralrobotmanipulation} & General Manipulation & PyBullet \cite{panerati2021learningflygym} & Multimodal prompts combining text and visuals for complex tasks. \\
\hline
VLABench \cite{zhang2024vlabenchlargescalebenchmarklanguageconditioned} & General Manipulation & MuJoCo Simulation \cite{mujoco} & Large-scale benchmark evaluating VLA models on complex long-horizon tasks requiring world knowledge. \\
\hline
DROID \cite{khazatsky2025droidlargescaleinthewildrobot} & General Manipulation & Real-world (Franka Panda) \cite{frankapanda} & Distributed dataset enhancing policy generalization across diverse real-world scenes and tasks. \\
\hline
RoboCasa \cite{nasiriany2024robocasalargescalesimulationeveryday} & Household task simulation & MuJoCo \cite{mujoco} Simulation & Large-scale benchmark with 100+ tasks and diverse kitchen scenes using procedural generation for generalist robot training. \\
\hline
VirtualHome \cite{puig2018virtualhome}& General Planning & Unity3D & Simulates complex household activities via programs; ideal for testing code-generating agents and multi-agent vulnerabilities. \\
\hline
Code as Policies \cite{liang2023code}& Code Generation Security & Real \& Sim & Generates Python policy code from language; its code-generation capability is a primary vector for security analysis. \\
\hline
ProgPrompt \cite{singh2022progpromptgeneratingsituatedrobot} & Code Generation Security & Real \& Sim (VirtualHome \cite{puig2018virtualhome}) & Generates situated task plans as programs; evaluated in complex environments like VirtualHome. \\
\hline
Instruct2Act \cite{huang2023instruct2actmappingmultimodalityinstructions} & Code Generation Security & Real \& Sim & Maps multi-modal instructions to executable Python programs for manipulation tasks. \\
\hline
VoxPoser \cite{huang2023voxposercomposable3dvalue} & Code Generation Security & Real \& Sim (RLBench \cite{james2020rlbench}) & Generates code to create 3D value maps for manipulation, enabling zero-shot generalization. \\
\hline
\end{tabular}
\label{tab:benchmarks}
\end{table*}

\section{Discussion}
\subsection{Environment Context is Critical}  
Unlike most LLM applications that primarily rely on text alignment, robotic systems inherently involve physical interactions. LLMs often generate infeasible plans for embodied agents due to lack of real-world knowledge, as their planning uses only internal knowledge that is not grounded in the physical world \cite{zhang2025efficientllmgroundingembodied}. LLMs often stumble when tackling basic physical reasoning or executing robotics tasks due to a lack of direct experience with the physical nuances of the real world \cite{liu2024groundinglargelanguagemodels}. This grounding deficit manifests as a critical misalignment between linguistic outputs and physical actions, where embodied LLMs may verbally refuse malicious requests while paradoxically executing the declined actions \cite{zhang2024badrobot}. Disembodied AI systems notoriously lack intuitive understanding of physical consequences and basic physics, often making nonsensical errors when confronted with situations requiring real-world understanding \cite{liu2025aligningcyberspacephysical}. This fundamental difference makes both attacks and defenses strongly dependent on the surrounding physical context, while current safety research remains largely focused on text-based alignment.

\subsection{Traditional defenses are not effective for LLM-controlled robotics.} Many current defense mechanisms are designed for text-only settings and fail to account for the embodiment gap and context-dependent physical risks in robotics. In embodied systems, preventing ``harmful text'' is not sufficient. Attacks must be prevented from yielding syntactically valid, physically executable commands that can cause real-world damage, which is explicitly highlighted in this survey's analysis for jailbreaking attacks~\cite{zhang2024badrobot, lu2025poexpolicyexecutablejailbreak, robey2024jailbreakingllmcontrolledrobots}. Moreover, harmfulness is contextual, undermining traditional, content-only safety filters that lack situational grounding~\cite{wu2025vulnerabilityllmvlmcontrolledrobotics}. Even within robotics-aware defenses, current approaches are siloed across layers: formal safety specifications provide guarantees only at the symbolic level and do not extend to continuous robot dynamics, while low-level control safeguards (e.g., CBFs) are decoupled from high-level LLM reasoning; together these gaps leave end-to-end assurance incomplete~\cite{ravichandran2025safetyguardrailsllmenabledrobots, 10611447, khan2025safetyawaretaskplanning}. Finally, assurance practices remain limited: evaluations often rely on static metrics and controlled simulations that do not capture long-horizon, multimodal uncertainty characteristic of real deployments, further evidencing why traditional defenses alone are insufficient\cite{Zhang_Kong_Dewitt_Bräunl_Hong_2025}.

\subsection{Research on Multimodal LLMs is demanded}
Multimodal LLMs expand the overall attack surface, as they involve multiple input modalities including text, images, audio, and sensor data \cite{chiu2025doisayi, wang2025alignenoughmultimodaluniversal,chiu2025say}. Existing research on the security and robustness of these systems remains limited, suggesting that dedicated efforts are required in this field. 

\subsection{Distinguishing Edge Cases and Comparing Mitigation Strategies}

While jailbreaking, backdoor, and prompt injection attacks all exploit the LLM–robot interface, their operational mechanisms, triggers, and mitigation pathways differ significantly. 
\textit{Jailbreaking attacks} represent inference-time manipulation, where adversaries coerce the planner to generate unsafe but syntactically valid action sequences. The defining edge case is \textit{policy executability}~\cite{zhang2024badrobot,lu2025poexpolicyexecutablejailbreak,robey2024jailbreakingllmcontrolledrobots,karnik2025embodiedredteamingauditing}, in which an attack is successful only if the malicious prompt results in commands compatible with the robot's APIs and constraints. Consequently, mitigation centers on runtime oversight and reasoning-layer defenses such as multi-LLM judges and formal safety guardrails (e.g., Safety Guardrails~\cite{ravichandran2025safetyguardrailsllmenabledrobots}, Safety Chip~\cite{10611447}), while syntax-checking components such as those in \textit{ROBOPAIR}~\cite{robey2024jailbreakingllmcontrolledrobots} serve as audit tools to detect executable syntax anomalies rather than enforce defenses at runtime. 

In contrast, \textit{Backdoor attacks} occur pre-deployment, embedding latent triggers that activate malicious behaviors only under specific contextual cues~\cite{wang2025trojanrobot}. These attacks persist silently until triggered, distinguishing them from jailbreaks that rely on active adversarial prompting. Such persistence may also affect fine-tuned or open-source models integrated into robotic pipelines. Mitigation therefore shifts toward provenance and auditing— including supply-chain integrity verification, training data validation, and periodic unlearning or fine-tuning audits. Because backdoors compromise the model itself, defenses must intervene before or during model integration rather than at inference time~\cite{Zhang_Kong_Dewitt_Bräunl_Hong_2025,yang2025ceeinferencetimejailbreakdefense}. 

\textit{Prompt Injection attacks}, finally, operate at the input and perception level, manipulating textual, visual, or sensor data streams in real time to redirect the model’s reasoning. Their edge cases arise from the multi-modal nature~\cite{wu2025vulnerabilityllmvlmcontrolledrobotics} of embodied systems—e.g., a benign textual command combined with a spoofed visual frame. Defenses accordingly prioritize input-layer hardening through schema filters, modality-specific sanitization, and network-level monitoring against man-in-the-middle interception~\cite{zhang2024safeembodaisafetyframeworkmobile}, aligning with mechanisms described under Prompt and Policy Hardening (F3) and Infrastructure-layer defenses in the taxonomy. 

Comparatively, jailbreaking mitigation emphasizes logical verification and runtime pruning; Backdoor defenses emphasize supply-chain assurance and model integrity, and Prompt Injection defenses emphasize input validation and multi-modal isolation. Together, these distinctions highlight that a unified ``one-size-fits-all'' defense is infeasible~\cite{khan2025safetyawaretaskplanning, Zhang_Kong_Dewitt_Bräunl_Hong_2025} - the mitigation surface must align precisely with the point of compromise across the perception, cognition, and control layers.

\section{Conclusion}
The rise of LLMs has given embodied intelligence the ability to understand complex instructions and to carry out reasoning across different contexts. At the same time, these advances expand the attack surface, creating new challenges for security and privacy that must be addressed. In this survey, we presented a systematic overview of attacks, defenses, and benchmarks in LLM-controlled robotics. We described major attack types, including jailbreaks, backdoor attacks, and prompt injection attacks, together with their subcategories. We also discussed defense strategies, divided into inference-time defenses and architectural/framework-based defenses. In addition, we reviewed benchmarks and datasets commonly used to test the robustness and safety of robotic systems under adversarial conditions, and we also provided a comprehensive integration roadmap for future work.

% \section*{Acknowledgments}
% This should be a simple paragraph before the References to thank those individuals and institutions who have supported your work on this article.

\balance
\bibliographystyle{IEEEtran}
\bibliography{references}

\end{document}